\documentclass[runningheads]{llncs}

\usepackage[mobile]{eccv}

\usepackage{eccvabbrv}

\usepackage{graphicx}
\usepackage{booktabs}

\usepackage[accsupp]{axessibility}  

\usepackage{amsmath}
\usepackage{amssymb}
\usepackage{caption}
\usepackage{multirow}
\usepackage{url}
\usepackage{xcolor}

\usepackage[pagebackref,breaklinks,colorlinks]{hyperref}

\usepackage{orcidlink}

\begin{document}

\title{LGM: Large Multi-View Gaussian Model for High-Resolution 3D Content Creation}

\titlerunning{Large Multi-View Gaussian Model}

\author{
Jiaxiang Tang\inst{1}\thanks{Work done while visiting S-Lab, Nanyang Technological University.} \and
Zhaoxi Chen\inst{2} \and
Xiaokang Chen\inst{1} \and 
Tengfei Wang\inst{3} \and
Gang Zeng\inst{1} \and
Ziwei Liu\inst{2}
}

\authorrunning{Tang et al.}

\institute{
National Key Lab of General AI, Peking University \and 
S-Lab, Nanyang Technological University \and 
Shanghai AI Lab
}

\maketitle

{
    \vspace{-0.5cm}
    \begin{center}
    \textbf{\url{https://me.kiui.moe/lgm}}
    \end{center}
    \vspace{-0.2cm}
    \centering
    \captionsetup{type=figure}
    \includegraphics[width=\textwidth]{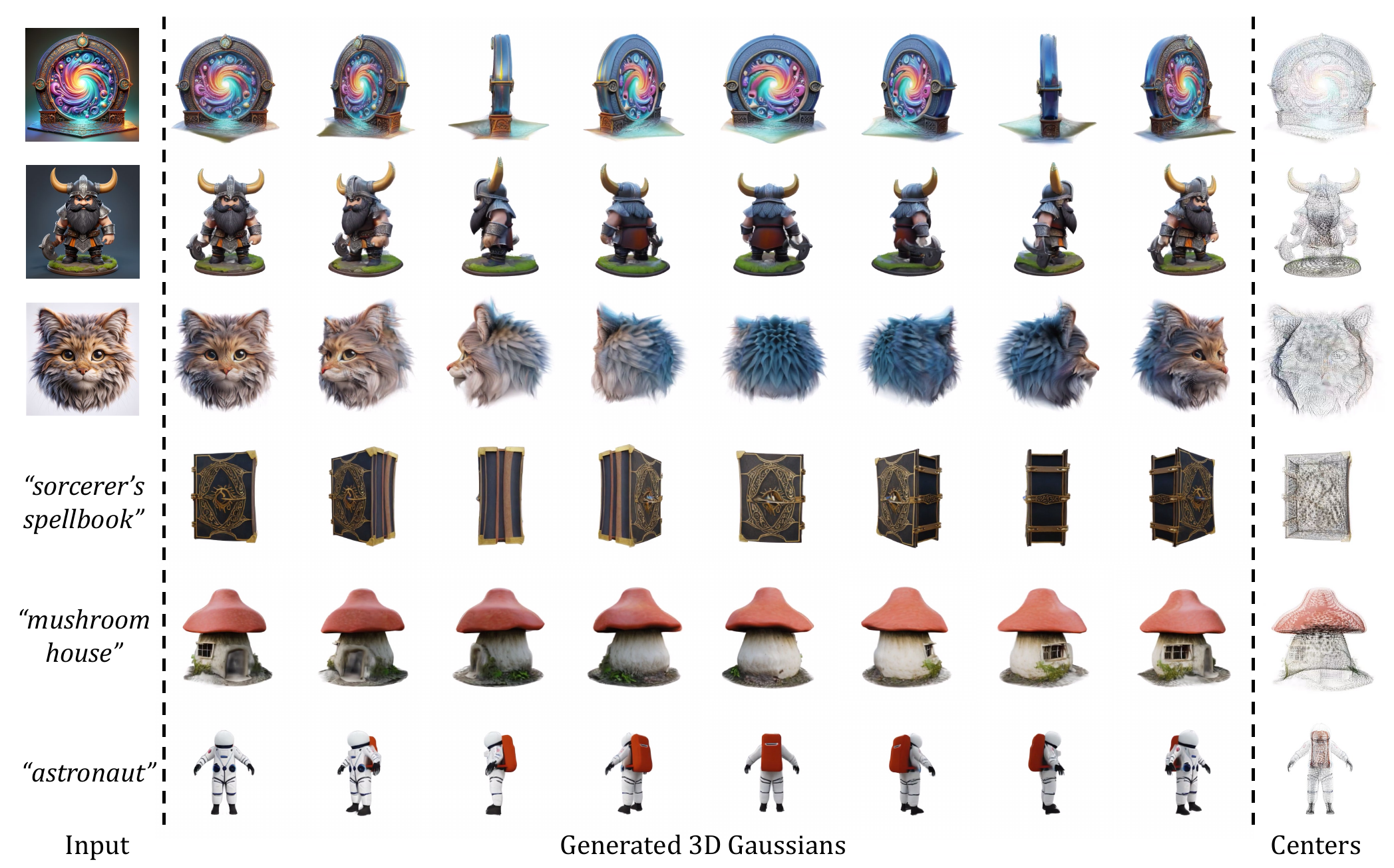}
    \captionof{figure}{
    Our method generates \textbf{high-resolution} 3D Gaussians in \textbf{5 seconds} from single-view images or texts.
    }
    \label{fig:teaser}
}

\begin{abstract}
3D content creation has achieved significant progress in terms of both quality and speed. 
Although current feed-forward models can produce 3D objects in seconds, their resolution is constrained by the intensive computation required during training. 
In this paper, we introduce \textbf{Large Multi-View Gaussian Model (LGM)}, a novel framework designed to generate high-resolution 3D models from text prompts or single-view images. 
Our key insights are two-fold: \textbf{1) 3D Representation:} We propose multi-view Gaussian features as an efficient yet powerful representation, which can then be fused together for differentiable rendering. 
\textbf{2) 3D Backbone:} We present an asymmetric U-Net as a high-throughput backbone operating on multi-view images, which can be produced from text or single-view image input by leveraging multi-view diffusion models. 
Extensive experiments demonstrate the high fidelity and efficiency of our approach. 
Notably, we maintain the fast speed to generate 3D objects within 5 seconds while boosting the training resolution to 512, thereby achieving high-resolution 3D content generation.

\end{abstract}

\section{Introduction}
\label{sec:intro}
Automatic 3D content creation has great potential in numerous fields such as digital games, virtual reality, and films. 
The fundamental techniques, like image-to-3D and text-to-3D, provide significant benefits by remarkably decreasing the requirement for manual labor among professional 3D artists, enabling those without expertise to participate in 3D asset creation.

Previous research on 3D generation has predominantly focused on score distillation sampling (SDS)~\cite{poole2022dreamfusion,lin2023magic3d,liu2023zero,tang2023dreamgaussian} to lift 2D diffusion priors into 3D. 
These optimization-based methods can create highly detailed 3D objects from text or single-view image inputs, but they often face issues such as slow generation speed and limited diversity. 
Recent advancements have significantly decreased the time required to generate 3D objects using large reconstruction models from single-view or few-shot images~\cite{hong2023lrm,wang2023pflrm,xu2023dmv3d,li2023instant3d,weng2024single}. 
These methods utilize transformers to directly regress triplane-based~\cite{Chan2022} neural radiance fields (NeRF)~\cite{mildenhall2020nerf}. 
However, these methods cannot produce detailed textures and complicated geometry due to the low-resolution training.
We argue that their bottlenecks are \textbf{1)} inefficient 3D representation, and \textbf{2)} heavily parameterized 3D backbone. 
For instance, given a fixed compute budget, the triplane representation of LRM~\cite{hong2023lrm} is limited to the resolution of $32$, while the resolution of the rendered image is capped at $128$ due to the online volume rendering.
Despite this, these methods suffer from the computationally intensive transformer-based backbone, which also leads to a limited training resolution.

To address these challenges, we present a novel method to train a few-shot 3D reconstruction model without relying on triplane-based volume rendering or transformers~\cite{hong2023lrm}. 
Instead, our approach employs 3D Gaussian splatting~\cite{kerbl20233d} of which features are predicted by an asymmetric U-Net as a high-throughput backbone~\cite{ronneberger2015u,szymanowicz23splatter}. 
The motivation of this design is to achieve high-resolution 3D generation, which necessitates an expressive 3D representation and the ability to train at high resolutions. 
Gaussian splatting stands out for 1) the expressiveness of compactly representing a scene compared with a single triplane, and 2) rendering efficiency compared with heavy volume rendering, which facilitates high-resolution training.
However, it requires a sufficient number of 3D Gaussians to accurately represent detailed 3D information.
Inspired by splatter image~\cite{szymanowicz23splatter}, we found that U-Net is effective in generating a sufficient number of Gaussians from multiview pixels, which maintains the capacity for high-resolution training at the same time.
Note that, compared to previous methods~\cite{hong2023lrm,zou2023triplane}, our default model is capable of generating 3D models with up to $65,536$ Gaussians and can be trained at a resolution of $512$, while still maintaining the rapid generation speed of feed-forward regression models. 
As shown in Figure~\ref{fig:teaser}, our model supports both image-to-3D and text-to-3D tasks, capable of producing high-resolution, richly detailed 3D Gaussians in approximately 5 seconds.

Our method adopts a multi-view reconstruction setting similar to Instant3D~\cite{li2023instant3d}. 
In this process, the image and camera embedding from each input view are transformed into a feature map, which can be decoded and fused as a set of Gaussians.  
Differentiable rendering is applied to render novel views from the fused 3D Gaussians, allowing end-to-end image-level supervision in high resolution. 
To enhance information sharing across all input views, attention blocks are integrated into the deeper layers of the U-Net. 
This enables us to train our network on multi-view image datasets~\cite{deitke2023objaverse} using only regressing objectives.
During inference, our method leverages existing image or text to multi-view diffusion models~\cite{shi2023mvdream,wang2023imagedream,shi2023zero123plus,long2023wonder3d} to produce multi-view images as inputs for our Gaussian fusion network. 
To overcome the domain gap between multi-view images rendered from actual 3D objects and synthesized using diffusion models, we further propose two proper data augmentations for robust training.
Finally, considering the preference for polygonal meshes in downstream tasks, we design a general algorithm to convert generated 3D Gaussians to smooth and textured meshes.

In summary, our contributions are:
\begin{enumerate}
    \item We propose a novel framework to generate high-resolution 3D Gaussians by fusing information from multi-view images, which can be generated from text prompts or single-view images.
    \item We design an asymmetric U-Net based architecture for efficient end-to-end training with significantly higher resolution, investigate data augmentation techniques for robust training, and propose a general mesh extraction approach from 3D Gaussians.
    \item Extensive experiments demonstrate the superior quality, resolution, and efficiency of our method in both text-to-3D and image-to-3D tasks.
\end{enumerate}

\section{Related Work}
\label{sec:rel}
\textbf{High-Resolution 3D Generation.} 
Current approaches for generating high-fidelity 3D models mostly rely on SDS-based optimization techniques. 
It requires both an expressive 3D representation and high-resolution supervision to effectively distill detailed information from 2D diffusion models into 3D. 
Due to the significant memory consumption associated with high-resolution rendering of NeRF, Magic3D~\cite{lin2023magic3d} first converts NeRF to DMTet~\cite{shen2021dmtet} and subsequently trains a second stage for finer resolution refinement. 
The hybrid representation of DMTet geometry and hash grid~\cite{mueller2022instant} textures enables the capture of high-quality 3D information, which can be efficiently rendered using differentiable rasterization~\cite{Laine2020diffrast}.
Fantasia3D~\cite{chen2023fantasia3d} explores to directly train DMTet with disentangled geometry and appearance generation. 
Subsequent studies~\cite{tsalicoglou2023textmesh,tang2023dreamgaussian,wang2023prolificdreamer,li2023focaldreamer,chen2023it3d,li2023sweetdreamer} also employ a similar mesh-based stage, enabling high-resolution supervision for enhanced detail.
Another promising 3D representation is Gaussian splatting~\cite{kerbl20233d} for its expressiveness and efficient rendering capabilities.
Nonetheless, achieving rich details with this method necessitates appropriate initialization and careful densification during optimization~\cite{chen2023gsgen,yi2023gaussiandreamer}. 
In contrast, our work investigates a feed-forward approach to directly generate a sufficient number of 3D Gaussians.

\noindent\textbf{Efficient 3D Generation.}
In contrast to SDS-based optimization methods, feed-forward 3D native methods are able to generate 3D assets within seconds after training on large-scale 3D datasets~\cite{deitke2023objaverse,deitke2023objaversexl}.
Some works attempt to train text-conditioned diffusion models on 3D representations such as point clouds and volumes~\cite{nichol2022point,jun2023shap,liu2023meshdiffusion,muller2023diffrf,chen2023primdiffusion,cao2023large,chen2023single,wang2023rodin,zhao2023michelangelo,yariv2023mosaic}.
However, these methods either cannot generalize well to large datasets or only produce low-quality 3D assets with simple textures.
Recently, LRM~\cite{hong2023lrm} first shows that a regression model can be trained to robustly predict NeRF from a single-view image in just 5 seconds, which can be further exported to meshes.
Instant3D~\cite{li2023instant3d} trains a text to multi-view images diffusion model and a multi-view LRM to perform fast and diverse text-to-3D generation.
The following works extend LRM to predict poses given multi-view images~\cite{wang2023pflrm}, combine with diffusion~\cite{xu2023dmv3d}, and specialize on human data~\cite{weng2024single}.
These feed-forward models can be trained with simple regression objectives and significantly accelerate the speed of 3D object generation.
However, their triplane NeRF-based representation is restricted to a relatively low resolution and limits the final generation fidelity.
Our model instead seeks to train a high-fidelity feed-forward model using Gaussian splatting and U-Net.

\noindent\textbf{Gaussian Splatting for Generation.}
We specifically discuss recent methods in generation tasks using Gaussian splatting~\cite{chen2024gssurvey,chen2023gaussianeditor,xu2024agg,ren2023dreamgaussian4d,ling2023align}. 
DreamGaussian~\cite{tang2023dreamgaussian} first combines 3D Gaussians with SDS-based optimization approaches to decrease generation time. 
GSGen~\cite{chen2023gsgen} and GaussianDreamer~\cite{yi2023gaussiandreamer} explore various densification and initialization strategies for text to 3D Gaussians generation. 
Despite the acceleration achieved, generating high-fidelity 3D Gaussians using these optimization-based methods still requires several minutes. 
TriplaneGaussian~\cite{zou2023triplane} introduces Gaussian splatting into the framework of LRM. 
This method starts by predicting Gaussian centers as point clouds and then projects them onto a triplane for other features. 
Nonetheless, the number of Gaussians and the resolution of the triplane are still limited, affecting the quality of the generated Gaussians. 
Splatter image~\cite{szymanowicz23splatter} proposes to predict 3D Gaussians as pixels on the output feature map using U-Net from single-view images. 
This approach mainly focuses on single-view or two-view scenarios, limiting its generalization to large-scale datasets. 
Similarly, PixelSplat~\cite{charatan2023pixelsplat} predicts Gaussian parameters for each pixel of two posed images from scene datasets.
We design a 4-view reconstruction model combined with existing multi-view diffusion models for general text or image to high-fidelity 3D object generation.

\section{Large Multi-View Gaussian Model}
\label{sec:method}
We first provide the background information on Gaussian splatting and multi-view diffusion models (Section~\ref{sec:bg}).
Then we introduce our high-resolution 3D content generation framework (Section~\ref{sec:pipe}), where the core part is an asymmetric U-Net backbone to predict and fuse 3D Gaussians from multi-view images (Section~\ref{sec:model}).
We design careful data augmentation and training pipeline to enhance robustness and stability (Section~\ref{sec:train}).
Finally, we describe an effective method for smooth textured mesh extraction from the generated 3D Gaussians (Section~\ref{sec:mesh}).

\subsection{Preliminaries}
\label{sec:bg}

\subsubsection{Gaussian Splatting.}

As introduced in~\cite{kerbl20233d}, Gaussian splatting employs a collection of 3D Gaussians to represent 3D data. 
Specifically, each Gaussian is defined by a center $\mathbf{x} \in \mathbb R^3$, a scaling factor $\mathbf{s} \in \mathbb R^3$, and a rotation quaternion $\mathbf{q} \in \mathbb R^4$. 
Additionally, an opacity value $\alpha \in \mathbb R$ and a color feature $\mathbf{c} \in \mathbb R^C$ are maintained for rendering, where spherical harmonics can be used to model view-dependent effects.
These parameters can be collectively denoted by ${\Theta}$, with ${\Theta}_i = \{\mathbf{x}_i, \mathbf{s}_i, \mathbf{q}_i, \alpha_i, \mathbf{c}_i\}$ representing the parameters for the $i$-th Gaussian. 
Rendering of the 3D Gaussians involves projecting them onto the image plane as 2D Gaussians and performing alpha composition for each pixel in front-to-back depth order, thereby determining the final color and alpha. 

\subsubsection{Multi-View Diffusion Models.}

Original 2D diffusion models~\cite{rombach2022high,saharia2022photorealistic} primarily focus on generating single-view images and do not support 3D viewpoint manipulation. 
Recently, several methods~\cite{shi2023mvdream,wang2023imagedream,shi2023zero123plus,long2023wonder3d,li2023sweetdreamer} propose to fine-tune multi-view diffusion models on 3D datasets to incorporate camera poses as an additional input. These approaches enable the creation of multi-view images of the same object, either from a text prompt or a single-view image. However, due to the absence of an actual 3D model, inconsistencies may still occur across the generated views.

\begin{figure*}[t!]
    \centering
    \includegraphics[width=\textwidth]{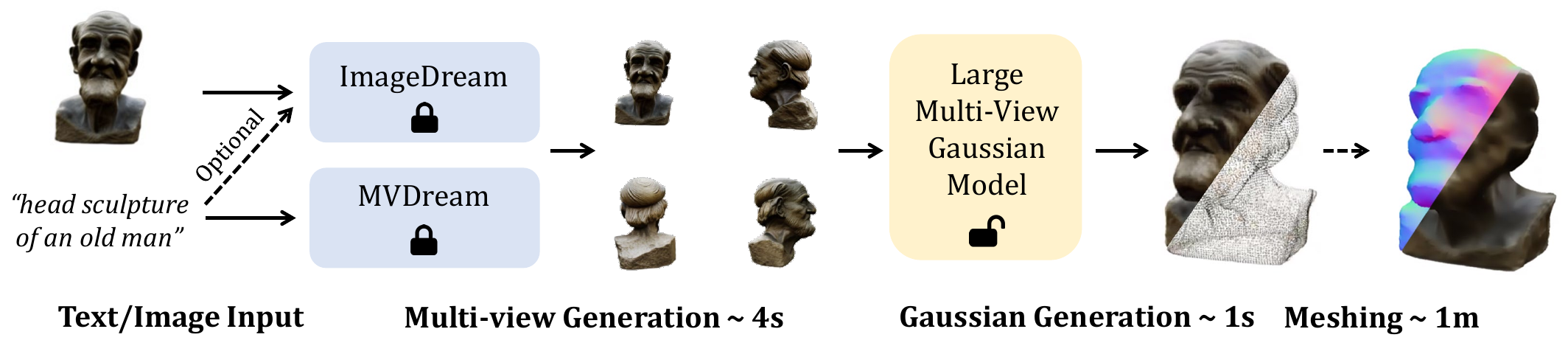}
    \caption{
    \textbf{Pipeline}. 
    Our model is trained to reconstruct 3D Gaussians from multi-view images, which can be synthesized by off-the-shelf models~\cite{shi2023mvdream,wang2023imagedream} at inference time from only text, or only image, or both input. Polygonal meshes can be extracted optionally.
    }
    \label{fig:pipe}
\end{figure*}

\begin{figure*}[t!]
    \centering
    \includegraphics[width=\textwidth]{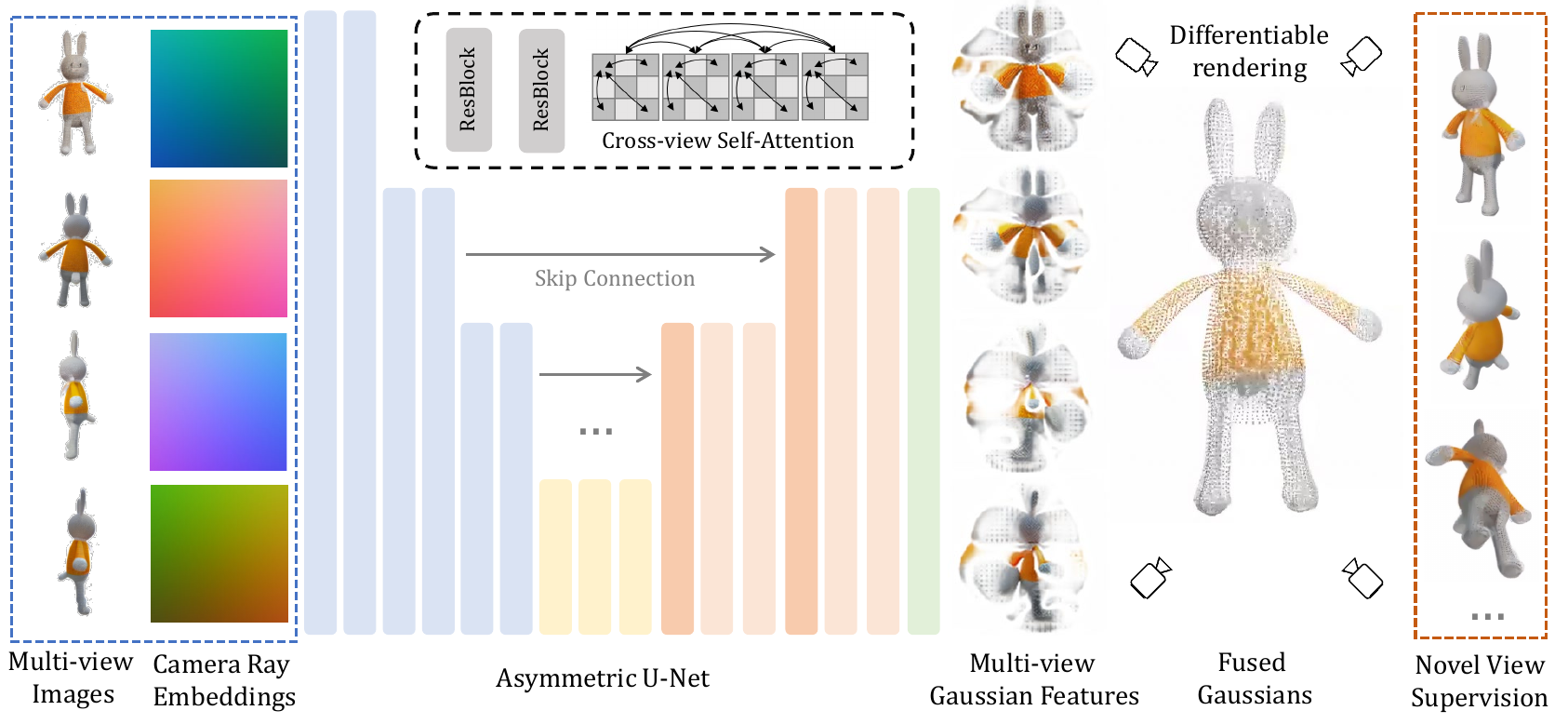}
    \caption{
    \textbf{Architecture of LGM}. 
    Our network adopts an asymmetric U-Net based architecture with cross-view self-attentions.
    We take four images with camera ray embeddings as the input, and output four feature maps which are interpreted and fused into 3D Gaussians.
    The Gaussians are then rendered at novel views and supervised with ground truth images.
    }
    \label{fig:method}
\end{figure*}

\subsection{Overall Framework}
\label{sec:pipe}

As illustrated in Figure~\ref{fig:pipe}, we adopt a two-step 3D generation pipeline at inference.
Firstly, we take advantage of off-the-shelf text or image to multi-view diffusion models to generate multi-view images.
Specifically, we adopt MVDream~\cite{shi2023mvdream} for text input and ImageDream~\cite{wang2023imagedream} for image (and optionally text) input.
Both models are designed to generate multi-view images at four orthogonal azimuths and a fixed elevation.
In the second step, we use a U-Net based model to predict 3D Gaussians from these sparse view images.
Specifically, our model is trained to take four images with camera pose embeddings as input and predict four sets of Gaussians, which are fused to form the final 3D Gaussians.
The generated Gaussians can be optionally converted to polygonal meshes using an extra conversion step, which is more suitable for downstream tasks.

\subsection{Asymmetric U-Net for 3D Gaussians}
\label{sec:model}

At the core of our framework is an asymmetric U-Net to predict and fuse Gaussians from multi-view images.
The network architecture is shown in Figure~\ref{fig:method}.
We take four images and corresponding camera poses as the input.
Following previous works~\cite{xu2023dmv3d}, we use the Plücker ray embedding to densely encode the camera poses.
The RGB value and ray embedding are concatenated into a 9-channel feature map as the input to the first layer:
\begin{equation}
    \mathbf{f}_i = \{\mathbf{c}_i, \mathbf{o}_i \times \mathbf{d}_i, \mathbf{d}_i\}
\end{equation}
where $\mathbf{f}_i$ is the input feature for pixel $i$, $\mathbf{c}_i$ is the RGB value, $\mathbf{d}_i$ is the ray direction, and $\mathbf{o}_i$ is the ray origin.

The U-Net is built with residual layers~\cite{he2016deep} and self-attention layers~\cite{vaswani2017attention} similar to previous works~\cite{szymanowicz23splatter,ho2020denoising,metzer2022latent}.
We only add self-attention at deeper layers where the feature map resolution is down-sampled to save memory.
To propagate information across multiple views, we flatten the four image features and concatenate them before applying self-attention, similar to previous multi-view diffusion models~\cite{shi2023mvdream,wang2023imagedream}.

Each pixel of the output feature map is treated as a 3D Gaussian inspired by splatter image~\cite{szymanowicz23splatter}.
Differently, our U-Net is designed to be asymmetric with a smaller output resolution compared to input, which allows us to use higher resolution input images and limit the number of output Gaussians.
We discard the depth prediction required by explicit ray-wise camera projection in~\cite{szymanowicz23splatter}.
The output feature map contains 14 channels corresponding to the original attributes of each Gaussian ${\Theta}_i$.
To stabilize the training, we choose some different activation functions compared to the original Gaussian Splatting~\cite{kerbl20233d}.
We clamp the predicted positions $\mathbf{x}_i$ into $[-1, 1]^3$, and multiply the softplus-activated scales $\mathbf{s}_i$ with $0.1$, such that the generated Gaussians at the beginning of training is close to the scene center.
For each input view, the output feature map is transformed into a set of Gaussians.
We simply concatenate these Gaussians from all four views as the final 3D Gaussians, which are used to render images at novel views for supervision.


\subsection{Robust Training}
\label{sec:train}

\subsubsection{Data Augmentation.}
We use multi-view images rendered from the Objaverse~\cite{deitke2023objaverse} dataset for training.
However, at inference, we use synthesized multi-view images by diffusion models~\cite{shi2023mvdream,wang2023imagedream}.
To mitigate the domain gap between these different multi-view images, we design two types of data augmentation for more robust training.

\noindent \textit{Grid Distortion.}
Synthesizing 3D consistent multi-view images using 2D diffusion models has been explored by many works~\cite{liu2023syncdreamer,shi2023mvdream,wang2023imagedream,shi2023zero123plus}.
However, since there is no underlying 3D representation, the generated multi-view images often suffer from subtle inconsistency across different views.
We try to simulate such inconsistency using grid distortion.
Except for the first input view, which is usually the front reference view, the other three input views are randomly distorted with a random grid during training.
This makes the model more robust to inconsistent multi-view input images.

\noindent \textit{Orbital Camera Jitter.}
Another problem is that the synthesized multi-view images may not accurately follow the given camera poses.
Following~\cite{hong2023lrm}, we always normalized the camera poses at each training step such that the first view's camera pose is fixed.
We therefore apply camera jitter to the last three input views during training.
Specifically, we randomly rotate the camera pose orbiting the scene center so the model is more tolerant to inaccurate camera poses and ray embeddings.

\subsubsection{Loss Function.}

To supervise the concatenated Gaussians, we use the differentiable renderer implementation from~\cite{kerbl20233d} to render them.
At each training step, we render the RGB image and alpha image of eight views, including four input views and four novel views.
Following~\cite{hong2023lrm}, we apply mean square error loss and VGG-based LPIPS loss~\cite{zhang2018perceptual} to the RGB image:
\begin{equation}
    \mathcal L_\text{rgb} = \mathcal L_\text{MSE} (I_\text{rgb}, I^\text{GT}_\text{rgb}) + \lambda\mathcal L_\text{LPIPS} (I_\text{rgb}, I^\text{GT}_\text{rgb})
\end{equation}
We further apply mean square error loss on the alpha image for faster convergence of the shape:
\begin{equation}
    \mathcal L_\alpha = \mathcal L_\text{MSE} (I_\alpha, I^\text{GT}_\alpha)
\end{equation}


\begin{figure*}[t!]
    \centering
    \includegraphics[width=\textwidth]{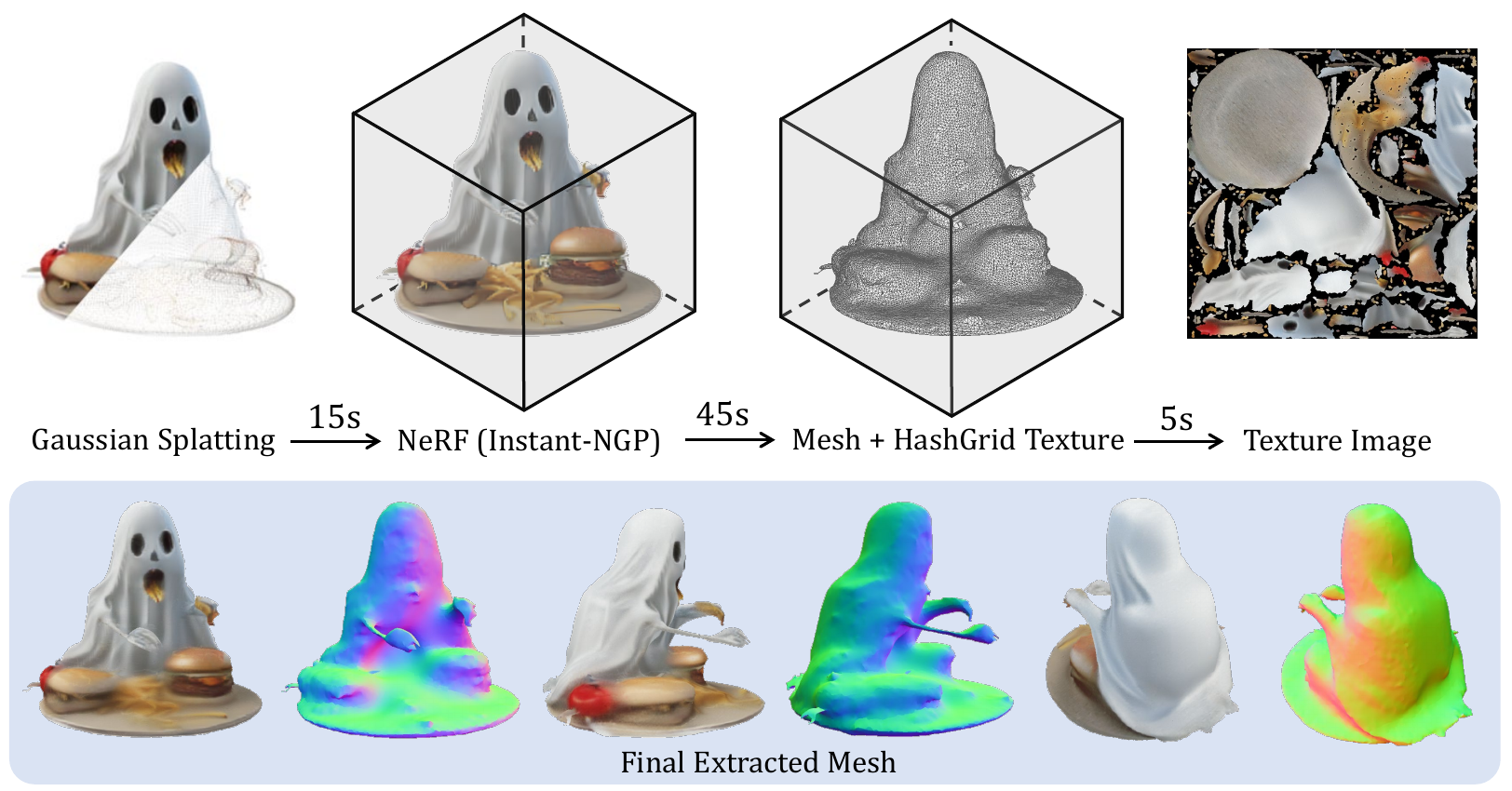}
    \caption{
    \textbf{Mesh Extraction Pipeline}. 
    We implement an efficient pipeline to convert the 3D Gaussians into smooth and textured meshes.
    }
    \label{fig:meshing}
\end{figure*}

\subsection{Mesh Extraction}
\label{sec:mesh}

Since polygonal meshes are still the most widely used 3D representation in downstream tasks, we hope to further extract meshes from our generated Gaussians.
Previous works~\cite{tang2023dreamgaussian} have tried to directly convert the opacity value of 3D Gaussians into an occupancy field for mesh extraction.
However, we find this method dependent on aggressive densification during the optimization of 3D Gaussians to produce smooth occupancy field.
On the contrary, the generated Gaussians in our method are usually sparse and cannot produce a suitable occupancy field, leading to an unsatisfactory surface with visible holes.

Instead, we propose a more general mesh extraction pipeline from 3D Gaussians as illustrated in Figure~\ref{fig:meshing}.
We first train an efficient NeRF~\cite{mueller2022instant} using the rendered images from 3D Gaussians on-the-fly, and then convert the NeRF to polygonal meshes~\cite{tang2022nerf2mesh}.
Specifically, we train two hash grids to reconstruct the geometry and appearance from Gaussian renderings. 
Marching Cubes~\cite{lorensen1998marching} is applied to extract a coarse mesh, which is then iteratively refined together with the appearance hash grid using differentiable rendering. 
Finally, we bake the appearance field onto the refined mesh to extract texture images.
For more details, please refer to the supplementary materials and NeRF2Mesh~\cite{tang2022nerf2mesh}.
With adequately optimized implementation, it takes only about 1 minute to perform this Gaussians to NeRF to mesh conversion.

\section{Experiments}
\label{sec:exp}

\subsection{Implementation Details}

\subsubsection{Datasets.}
We use a filtered subset of the Objaverse~\cite{deitke2023objaverse} dataset to train our model.
Since there are many low-quality 3D models (\textit{e.g.}, partial scans, missing textures) in the original Objaverse dataset, we filter the dataset by two empirical rules:
(1) We manually examine the captions and rendered images from Cap3D~\cite{luo2023scalable}, and curate a list of words that usually appears in bad models (\textit{e.g.}, `resembling', `debris', `frame'), which is then used to filter all models whose caption includes any of these words.
(2) We discard models with mostly white color after rendering, which usually indicates missing texture.
These lead to a final set of around 80K 3D objects. 
We render the RGBA image from 100 camera views at the resolution of $512\times 512$ for training and validation.

\subsubsection{Network Architecture.}
Our asymmetric U-Net model consists of 6 down blocks, 1 middle block, and 5 up blocks, with the input image at $256\times 256$ and output Gaussian feature map at $128 \times 128$.
We use 4 input views, so the number of output Gaussians is $128 \times 128 \times 4 = 65,536$.
The feature channels for all blocks are $[64, 128, 256, 512, 1024, 1024]$, $[1024]$ and $[1024, 1024, 512, 256, 128]$ respectively.
Each block contains a series of residual layers and an optional down-sample or up-sample layer.
For the last 3 of down blocks, the middle block, and the first 3 up blocks, we also insert cross-view self-attention layers after the residual layers.
The final feature maps are processed by a $1\times 1$ convolution layer to 14-channel pixel-wise Gaussian features.
Following previous works~\cite{szymanowicz23splatter,rombach2022high}, we adopt \texttt{Silu} activation and group normalization for the U-Net.

\begin{figure*}[ht!]
    \centering
    \includegraphics[width=\textwidth]{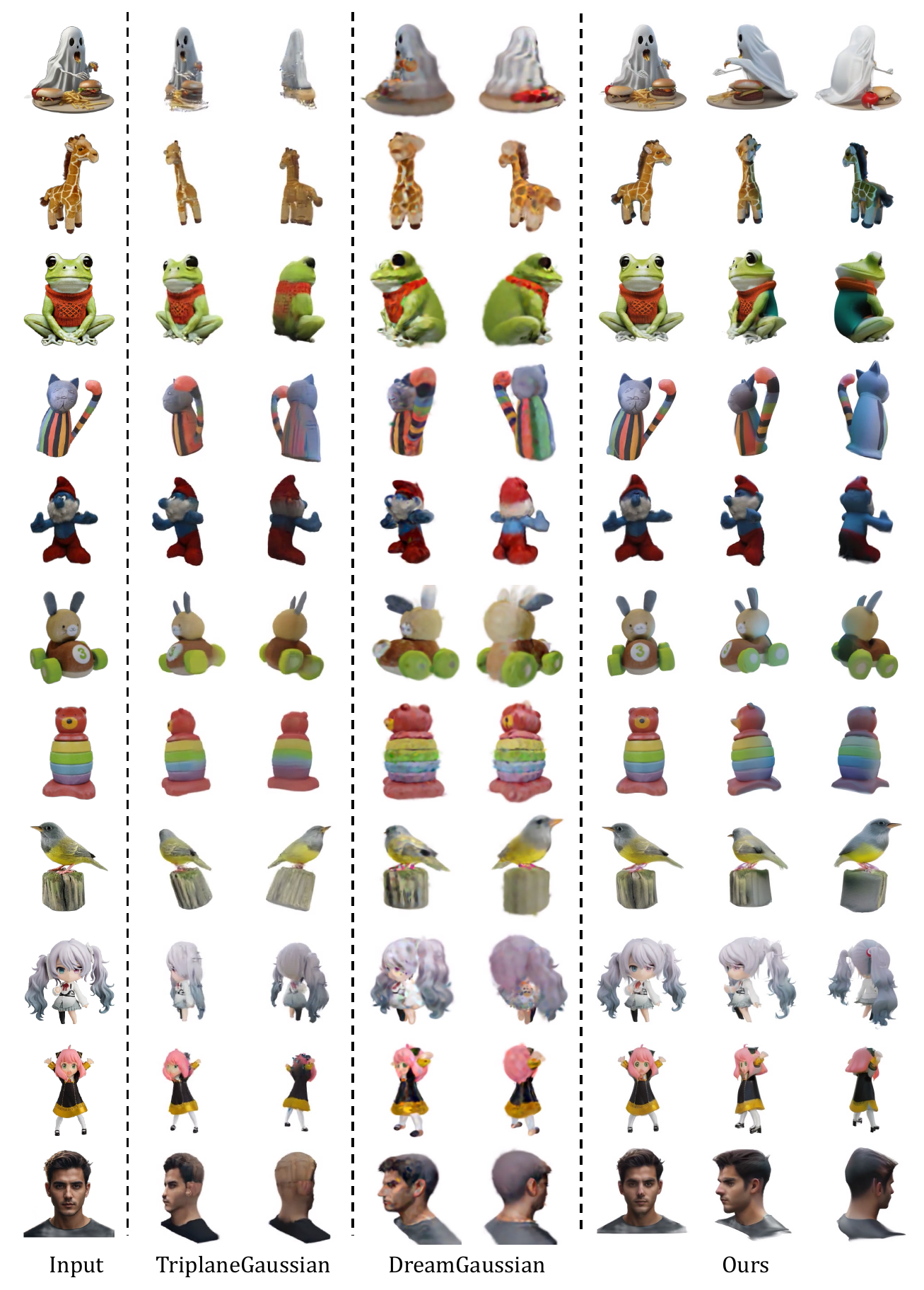}
    \caption{
    \textbf{Comparisons of generated 3D Gaussians for image-to-3D}. 
    Our method generates Gaussian splatting with better visual quality on various challenging images.
    }
    \label{fig:image_comp_gs}
\end{figure*}

\begin{figure*}[t!]
    \centering
    \includegraphics[width=\textwidth]{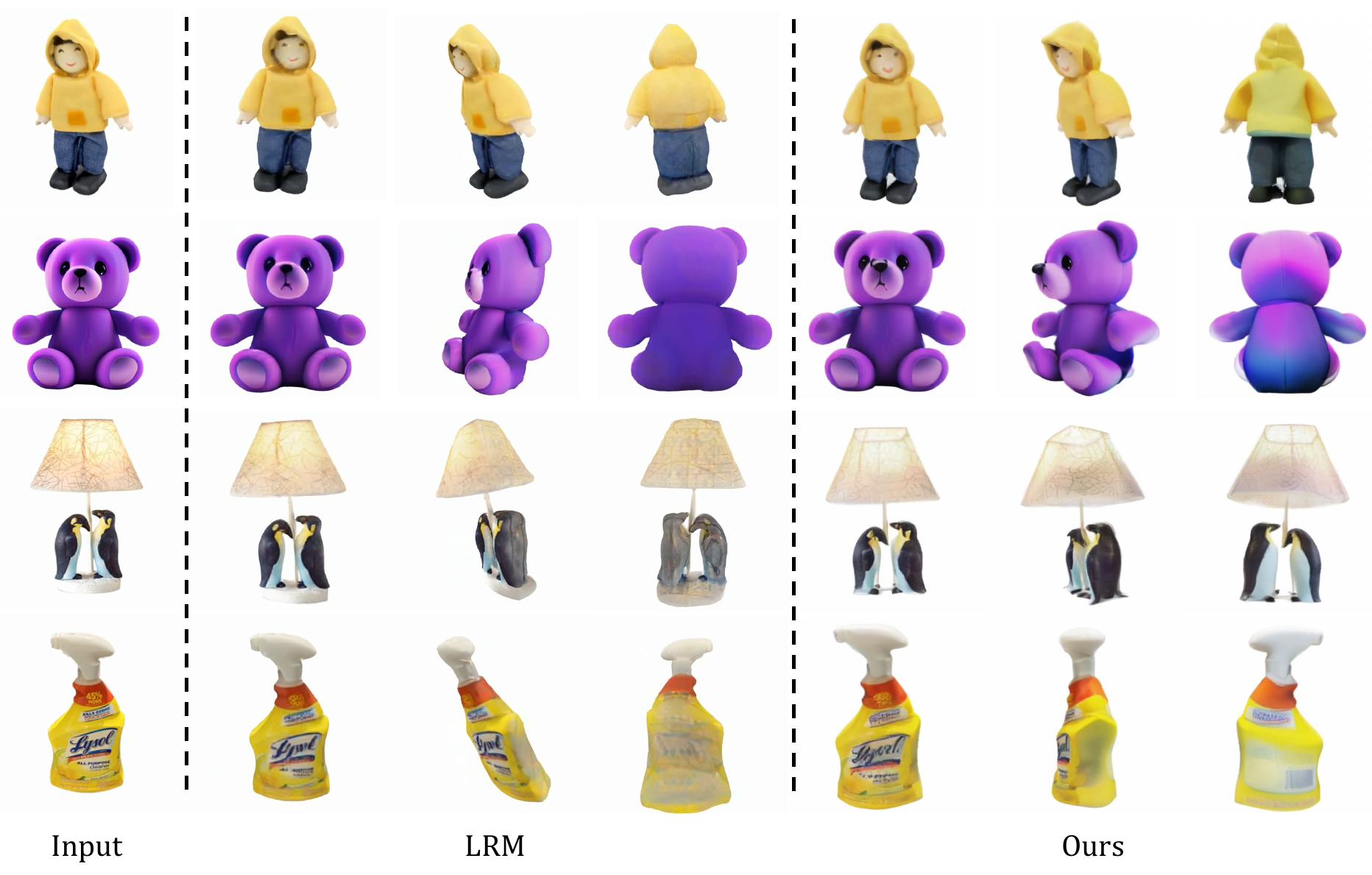}
    \caption{
    \textbf{Comparisons with LRM for image-to-3D}. 
    We compare our method with available results from LRM~\cite{hong2023lrm}.
    }
    \label{fig:image_comp_lrm}
\end{figure*}

\begin{figure*}[ht!]
    \centering
    \includegraphics[width=\textwidth]{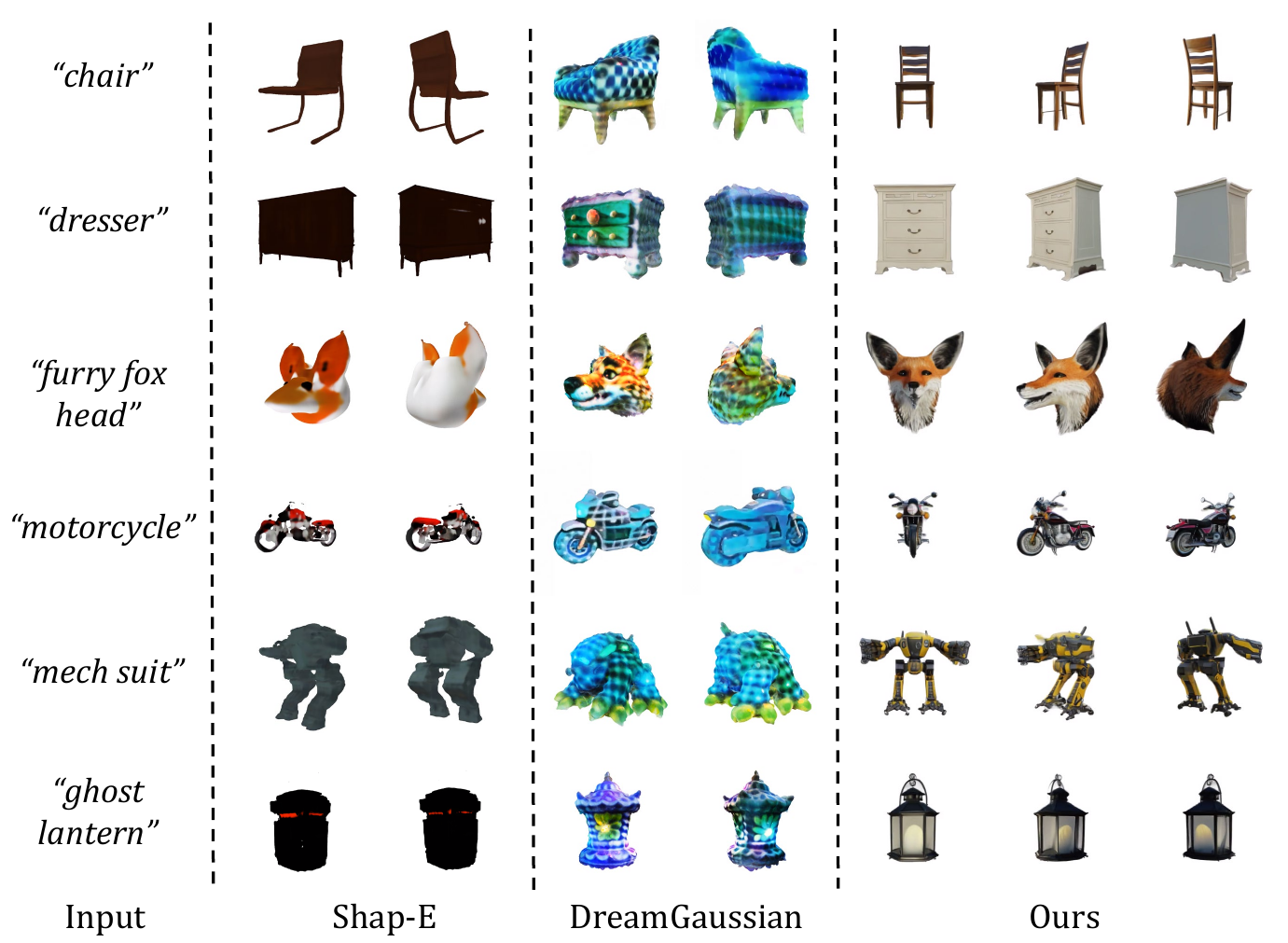}
    \caption{
    \textbf{Comparisons of generated 3D models for text-to-3D}. 
    Our method achieves better text alignment and visual quality.
    }
    \label{fig:text_comp}
\end{figure*}

\begin{figure*}[ht!]
    \centering
    \includegraphics[width=\textwidth]{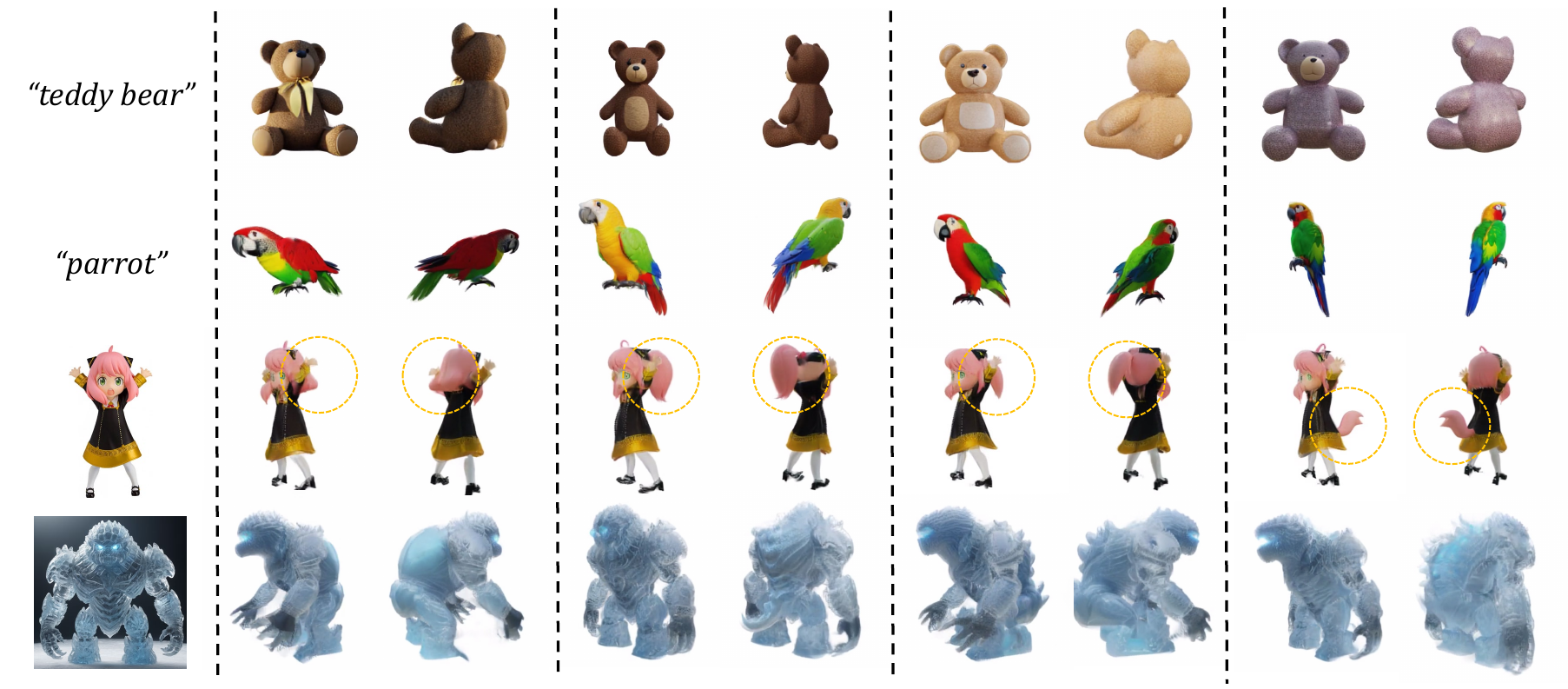}
    \caption{
    \textbf{Diversity of our 3D generation}.
    We can generate diverse 3D models given an ambiguous text description or single-view image.
    }
    \label{fig:diversity}
\end{figure*}

\subsubsection{Training.}
We train our model on 32 NVIDIA A100 (80G) GPUs for about 4 days.
A batch size of 8 for each GPU is used under \texttt{bfloat16} precision, leading to an effective batch size of 256.
For each batch, we randomly sample 8 camera views, with the first 4 views as the input, and all 8 views as the output for supervision.
Similar to LRM~\cite{hong2023lrm}, we transform the cameras of each batch such that the first input view is always the front view with an identity rotation matrix and fixed translation.
The input images are assumed to have a white background.
The output 3D Gaussians are rendered at $512 \times 512$ resolution for mean square error loss.
We resize the images to $256 \times 256$ for LPIPS loss to save memory.
The AdamW~\cite{loshchilov2017decoupled} optimizer is adopted with the learning rate of $4 \times 10^{-4}$, weight decay of $0.05$, and betas of $(0.9, 0.95)$. The learning rate is cosine annealed to $0$ during the training.
We clip the gradient with a maximum norm of $1.0$.
The probability for grid distortion and camera jitter is set to $50$\%.

\subsubsection{Inference.}
Our whole pipeline, including two multi-view diffusion models, takes only about 10 GB of GPU memory for inference, which is friendly for deployment.
For the multi-view diffusion models, we use a guidance scale of 5 for ImageDream~\cite{wang2023imagedream} and 7.5 for MVDream~\cite{shi2023mvdream} following the original paper.
The number of diffusion steps is set to 30 using the DDIM~\cite{song2020denoising} scheduler.
The camera elevation is fixed to 0, and azimuths to $[0, 90, 180, 270]$ degree for the four generated views.
For ImageDream~\cite{wang2023imagedream}, the text prompt is always left empty so the only input is a single-view image.
Since the images generated by MVDream may contain various backgrounds, we apply background removal~\cite{qin2020u2} and use white background.

\subsection{Qualitative Comparisons}

\subsubsection{Image-to-3D.}

We first compare against recent methods~\cite{zou2023triplane,tang2023dreamgaussian} that are capable of generating 3D Gaussians. 
Figure~\ref{fig:image_comp_gs} shows images rendered from the generated 3D Gaussians for comparison. 
The 3D Gaussians produced by our method have better visual quality and effectively preserve the content from the input view. 
Our high-resolution 3D Gaussians can be transformed into smooth textured meshes with minimal loss of quality in most cases.
We also compare our results against LRM~\cite{hong2023lrm} using the available videos from their website in Figure~\ref{fig:image_comp_lrm}.
Specifically, our multi-view setting successfully mitigates the issue of blurry back views and flat geometry, resulting in enhanced detail even in unseen views.

\subsubsection{Text-to-3D.}
We then compare with recent methods~\cite{jun2023shap,tang2023dreamgaussian} on text-to-3D tasks. 
We observe an enhanced quality and efficiency in our method, generating more realistic 3D objects, as illustrated in Figure~\ref{fig:text_comp}. 
Due to the multi-view diffusion models, our model is also free from multi-face problems.

\subsubsection{Diversity.}
Notably, our pipeline exhibits high diversity in 3D generation, owing to the capability of multi-view diffusion model~\cite{shi2023mvdream,wang2023imagedream}.
As shown in Figure~\ref{fig:diversity}, with different random seeds, we can generate a variety of feasible objects from the same ambiguous text prompt or single-view image.

\subsection{Quantitative Comparisons}

\begin{table*}[!t]
\begin{center}
\setlength{\tabcolsep}{15pt}
\begin{tabular}{l|c|c}
\hline
              & Image Consistency $\uparrow$  & Overall Quality $\uparrow$ \\
\hline
DreamGaussian~\cite{tang2023dreamgaussian} & 2.30 & 1.98  \\
TriplaneGaussian~\cite{zou2023triplane}    & 3.02 & 2.67  \\
LGM (Ours)                                       & \textbf{4.18} & \textbf{3.95}  \\
\hline
\end{tabular}

\end{center}
\caption{
\textbf{User Study} on the quality of generated 3D Gaussians for image-to-3D tasks. The rating is of scale 1-5, the higher the better.
}
\label{tab:userstudy}
\end{table*}

We majorly conduct a user study to quantitatively evaluate our image-to-3D Gaussians generation performance.
For a collection of 30 images, we render 360-degree rotating videos of the 3D Gaussians generated from DreamGaussain~\cite{tang2023dreamgaussian} (only the first stage), TriplaneGaussian~\cite{zou2023triplane}, and ours.
There are in total 90 videos for evaluation in our user study.
Each volunteer is shown 30 samples from mixed random methods, and asked to rate in two aspects: image consistency and overall model quality.
We collect results from 20 volunteers and get 600 valid scores in total.
As shown in Table~\ref{tab:userstudy}, our method is preferred as it aligns with the original image content and shows better overall quality.

\subsection{Ablation Study}

\begin{figure*}[t!]
    \centering
    \includegraphics[width=\textwidth]{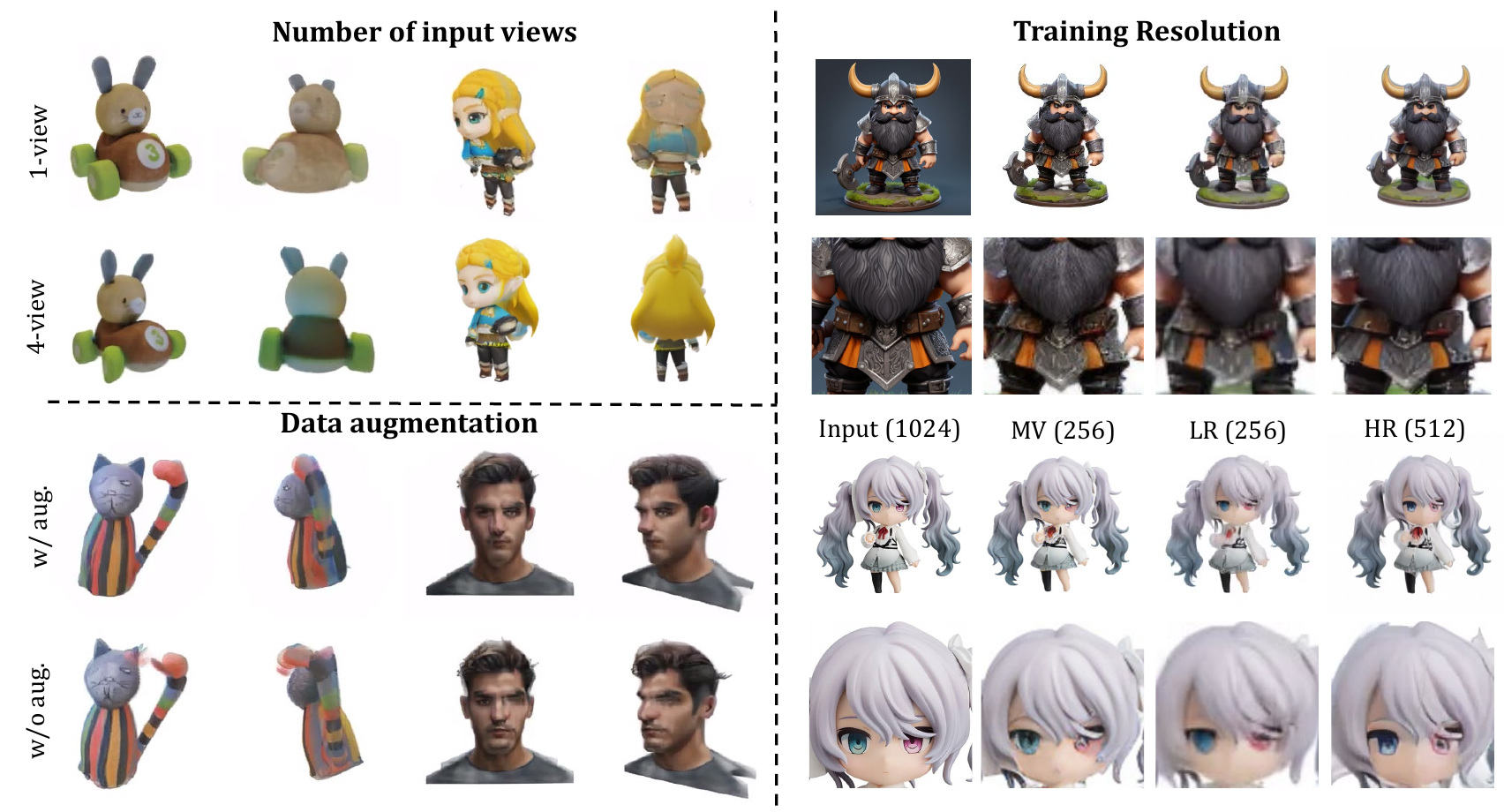}
    \caption{
    \textbf{Ablation Study}.
    We carry out ablation study on designs of our method.
    }
    \label{fig:ablation}
\end{figure*}

\subsubsection{Number of Views.}
We train an image-to-3D model with only one input views similar to splatter image~\cite{szymanowicz23splatter}, \textit{i.e.}, without the multi-view generation step.
The U-Net takes the single input view as input with self-attention, and outputs Gaussian features as in our multi-view model.
To compensate the number of Gaussians, we predict two Gaussians for each pixel of the output feature maps, leading to $128\times 128 \times 2 = 32,768$ Gaussians.
As illustrated in the top-left part of Figure~\ref{fig:ablation}, the single-view model can reconstruct faithful front-view, but fails to distinguish the back view and results in blurriness.
This is as expected since the regressive U-Net is more suitable for reconstruction tasks, and it's hard to generalize to large datasets in our experiments.

\subsubsection{Data Augmentation.}
We train a smaller model with or without applying data augmentation to validate its effectiveness.
Although we observe a lower training loss for the model without data augmentation, the domain gap during inference leads to more floaters and worse geometry as shown in the bottom-left part of Figure~\ref{fig:ablation}.
The model with data augmentation strategy can better correct the 3D inconsistency and inaccurate camera poses in the generated multi-view images.

\subsubsection{Training Resolution.}
Lastly, we train a model with a fewer number of Gaussians and smaller rendering resolution as in the right part of Figure~\ref{fig:ablation}.
We remove the last up block of the U-Net so the number of output Gaussians is $64 \times 64 \times 4 = 16,384$, and we render it at $256 \times 256$ for supervision.
The model can still converge and successfully reconstruct 3D Gaussians, but the details are worse compared to the $256 \times 256$ input multi-view images.
In contrast, our large resolution model at $512 \times 512$ can capture better details and generate Gaussians with higher resolution.

\subsection{Limitations}
Despite the promising results, our method still has some limitations. 
Since our model is essentially a multi-view reconstruction model, the 3D generation quality highly depends on the quality of four input views. 
However, current multi-view diffusion models~\cite{shi2023mvdream,wang2023imagedream} are far from perfect: 
(1) There can be 3D inconsistency which misleads the reconstruction model to generate floaters in the 3D Gaussians. 
(2) The resolution of synthesized multi-view images is restricted to $256 \times 256$, constraining our model to further improve resolution. 
(3) ImageDream~\cite{wang2023imagedream} also fails to handle input image with a large elevation angle. 
We expect these limitations can be mitigated with better multi-view diffusion models in future works.

\section{Conclusion}
In this work, we present a large multi-view Gaussian model for high-resolution 3D content generation. 
Our model, distinct from previous methods reliant on NeRF and transformers, employs Gaussian splatting and U-Net to address the challenges of high memory requirements and low-resolution training. 
Additionally, we explore data augmentation for better robustness, and introduce a mesh extraction algorithm for the generated 3D Gaussians. 
Our approach achieves both high-resolution and high-efficiency for 3D objects generation, proving its versatility and applicability in various contexts.

\subsubsection{Acknowledgements.}
This work is supported by the Sichuan Science and Technology Program (2023YFSY0008), China Tower-Peking University Joint Laboratory of Intelligent Society and Space Governance, National Natural Science Foundation of China (61632003, 61375022, 61403005), Grant SCITLAB-20017 of Intelligent Terminal Key Laboratory of SiChuan Province, Beijing Advanced Innovation Center for Intelligent Robots and Systems (2018IRS11), and PEK-SenseTime Joint Laboratory of Machine Vision. This project is also funded by the Ministry of Education, Singapore, under its MOE AcRF Tier 2 (MOE-T2EP20221-0012), NTU NAP, and under the RIE2020 Industry Alignment Fund – Industry Collaboration Projects (IAF-ICP) Funding Initiative, as well as cash and in-kind contribution from the industry partner(s).

%
%
\bibliographystyle{eccv_refstyle}
\bibliography{ref}

\clearpage

\appendix
\section{More Implementation Details}

\subsubsection{Datasets.}
The full list of words for filtering the Objaverse dataset is: `flying, mountain, trash, featuring, a set of, a small, numerous, square, collection, broken, group, ceiling, wall, various, elements, splatter, resembling, landscape, stair, silhouette, garbage, debris, room, preview, floor, grass, house, beam, white, background, building, cube, box, frame, roof, structure'.
The 100 camera views we use form a spiral path on the sphere surface. The camera radius is fixed to 1.5, and the field-of-view along the Y-axis is fixed to 49.1 degree.

\section{More Results}

\begin{figure*}[ht]
    \centering
    \includegraphics[width=\textwidth]{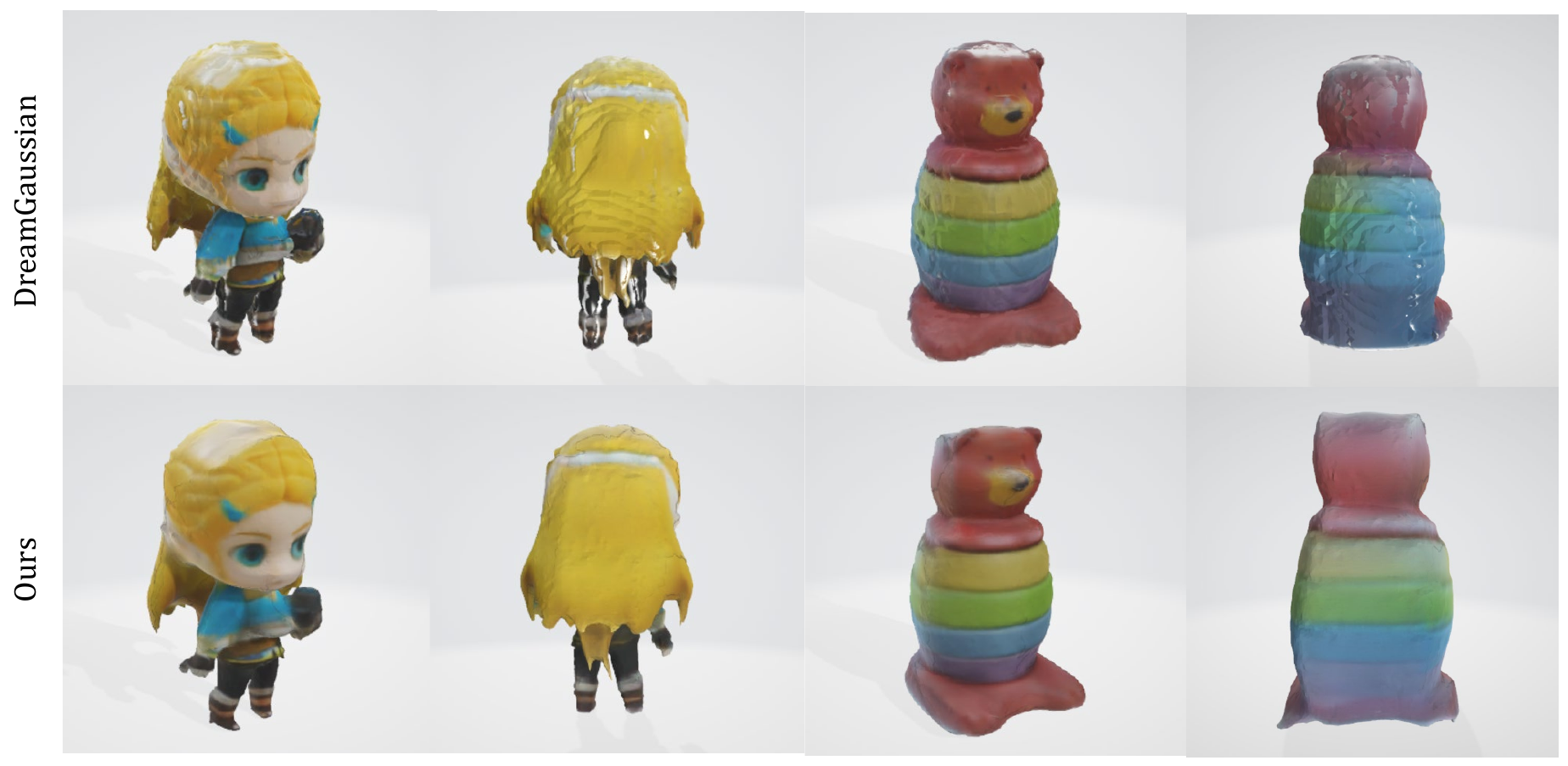}
    \caption{
    \textbf{Comparisons between different meshing method from Gaussians}.
    We compare our meshing method with DreamGaussian~\cite{tang2023dreamgaussian}.
    }
    \label{fig:comp_meshing}
\end{figure*}

\subsubsection{Different Meshing Method.}
Figure~\ref{fig:comp_meshing} presents a comparison between our meshing algorithm and the technique introduced in DreamGaussian~\cite{tang2023dreamgaussian}.
Our algorithm generates a smoother surface, which is advantageous for subsequent tasks such as relighting. 
Moreover, our method operates independently of the underlying 3D Gaussians, as it relies solely on the rendered images.

\begin{figure*}[ht]
    \centering
    \includegraphics[width=\textwidth]{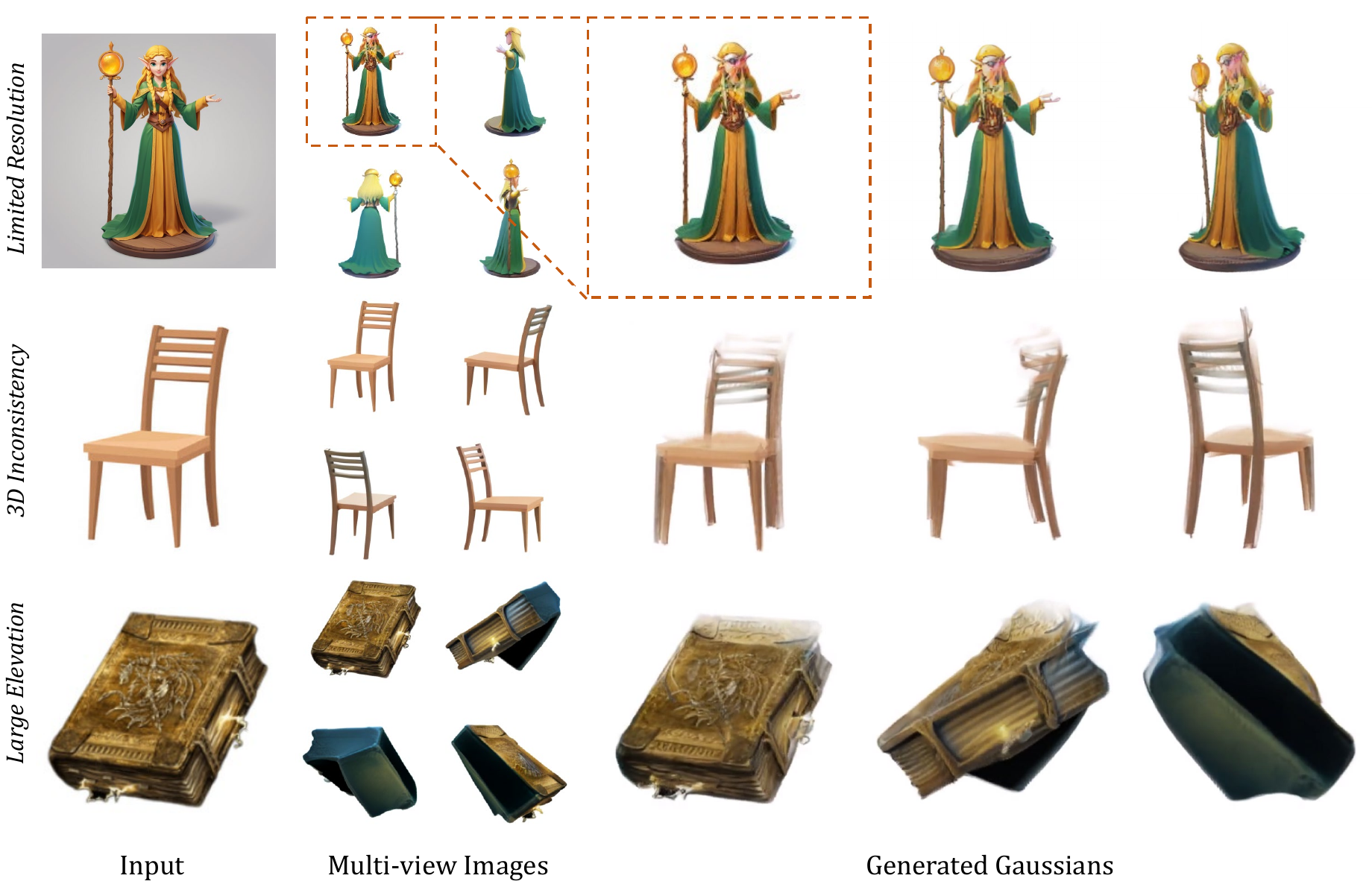}
    \caption{
    \textbf{Visualization of our limitations}.
    We show three major reasons for failure cases of our method.
    }
    \label{fig:limit}
\end{figure*}

\subsubsection{Limitations.}

We visualize failure cases of our method in Figure~\ref{fig:limit} to gain a deeper understanding of its weaknesses. 
As previously mentioned in the main paper, the primary causes of these failures stem from the flawed multi-view images produced in the initial step. 
The resolution of these multi-view images is limited to $256 \times 256$, which can diminish the quality of the input image. 
Despite implementing data augmentation during training to emulate 3D inconsistencies and attempting to bridge the domain gap, this approach still results in inaccuracies for slender structures, such as chairs. 
Additionally, ImageDream~\cite{wang2023imagedream} struggles with images that have significant elevation angle, occasionally producing images with a dark appearance.

\end{document}